\documentclass{article}

\usepackage{microtype}
\usepackage{graphicx}
\usepackage{subfigure}
\usepackage{booktabs} 
\usepackage{hyperref}

\usepackage[accepted]{ml4astro}

\usepackage{amsmath}
\usepackage{amssymb}
\usepackage{mathtools}
\usepackage{amsthm}

\usepackage[capitalize,noabbrev]{cleveref}

\theoremstyle{plain}

\theoremstyle{definition}

\theoremstyle{remark}

\usepackage[textsize=tiny]{todonotes}

\mlforastrotitlerunning{Set-based Implicit Likelihood Inference of Galaxy Cluster Mass}

\begin{document}

\twocolumn[

\mlforastrotitle{Set-based Implicit Likelihood Inference of Galaxy Cluster Mass}

\mlforastrosetsymbol{equal}{*}

\begin{mlforastroauthorlist}
\mlforastroauthor{Bonny Y. Wang}{cmu,cmuetc,ipmu,cd3}
\mlforastroauthor{Leander Thiele}{ipmu,cd3}

\end{mlforastroauthorlist}

\mlforastroaffiliation{cmu}{McWilliams Center for Cosmology and Astrophysics, Department of Physics, Carnegie Mellon University, Pittsburgh, PA 15213, USA}
\mlforastroaffiliation{cmuetc}{Entertainment Technology Center, Carnegie Mellon University, Pittsburgh, PA 15213, USA}
\mlforastroaffiliation{ipmu}{Kavli IPMU (WPI), UTIAS, The University of Tokyo, Japan}
\mlforastroaffiliation{cd3}{Center for Data-Driven Discovery, Kavli IPMU (WPI), Japan}

\mlforastrocorrespondingauthor{Bonny Y. Wang}{bonnyw@uchicago.edu}

\mlforastrokeywords{Machine Learning, Cosmology, Galaxy Cluster, ICML}

\vskip 0.3in
]

\printAffiliationsAndNotice{}

\begin{abstract}
We present a set-based machine learning framework that infers posterior distributions of galaxy cluster masses from projected galaxy dynamics. Our model combines Deep Sets and conditional normalizing flows to incorporate both positional and velocity information of member galaxies to predict residual corrections to the $M$–$\sigma$ relation for improved interpretability. Trained on the Uchuu-UniverseMachine simulation, our approach significantly reduces scatter and provides well-calibrated uncertainties across the full mass range compared to traditional dynamical estimates.

\end{abstract}

\section{Introduction}
\begin{figure}[!htb]
    \centering
    \includegraphics[width=0.98\linewidth]{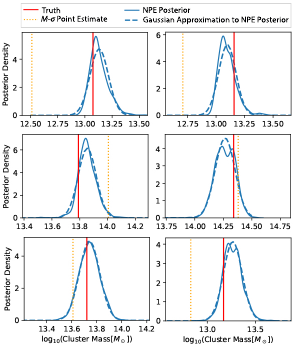}
    \caption{Posterior distributions of galaxy cluster mass estimates. The solid blue lines show posteriors predicted by the Neural Posterior Estimator (NPE), while the blue dashed lines indicate Gaussian approximations to these generated distributions. The red vertical lines mark the true cluster masses, and the orange dotted lines denote baseline $M$-$\sigma$ mass estimates. The left column presents examples of unimodal distributions, whereas the right column highlights bimodal cases.}
    \label{fig:dists}
\end{figure}

Galaxy clusters are powerful cosmological probes that constrain fundamental parameters of the Universe \citep{RevModPhys.77.207,2011ARA&A..49..409A,2016arXiv160407626D,2001ApJ...553..545H}. Given the current $S_8$ tension, accurately characterizing the cluster mass function is critical for testing and assessing the standard $\Lambda$CDM model.

Currently, we use a wide range of observational strategies for cluster mass estimation. These includes X-ray measurements of intra-cluster gas\citep[e.g.,][]{10.1093/mnras/stw2250,2017MNRAS.465..858G}, thermal Sunyaev-Zel’dovich (tSZ) effect, \citep[e.g.,][]{2021ApJS..253....3H}, and richness information \citep[e.g.,][]{2003ApJ...585..215Y}. More direct methods include weak gravitational lensing \citep[e.g.,][]{2014MNRAS.439...48A,2019MNRAS.482.1352M} and galaxy kinematics \citep[e.g.,][]{1933AcHPh...6..110Z,2014MNRAS.441.1513O,Geller_2013}, though these approaches are either limited by low signal-to-noise ratios, projection effects, or fail to fully capture baryonic modeling uncertainties.

The classic $M$–$\sigma$ relation provides a simple dynamical mass approximation based on the velocity dispersion of member galaxies. However, it relies on idealized assumptions that clusters are perfectly spherical, composed of equal-mass galaxies, and in a state of virial equilibrium. In reality, galaxy clusters are more dynamically complex and often far from equilibrium. To overcome these limitations, recent work has incorporated deep learning approaches \citep[e.g.,][]{Ntampaka2016, Ntampaka2019, Ho2019, Ho2021, Ho2022, Ho2023, Yan2020, KodiRamanah2021, Gupta2020, Gupta2021, Garuda2024} to model nonlinear patterns and extract information from high-dimensional galaxy dynamics. These previous methods often omit spatial and morphological information, rely mostly on point estimates, or compress the data. 

Here, we employ Deep Sets architecture that incorporates the full projected phase-space information of member galaxies, including both their velocities and positions, alongside global cluster morphology. This architecture is well suited for astronomical problems involving unordered collections with varying numbers of objects \citep[see e.g.,][]{Thiele2022, Wang2023, Jung2023, DeSanti2023}.

Our method fits within the broader context of simulation-based inference (SBI) \citep{Cranmer2020}. We specifically build on recent advances in neural posterior estimation (NPE) \citep{Greenberg2019, Lueckmann2027, Papamakarios2016} by using neural spline flows \citep{Durkan2019, Dolatabadi2020}. This enables accurate uncertainty quantification by producing expressive posterior distributions for galaxy cluster masses. As shown in Figure~\ref{fig:dists}, the predicted posteriors flexibly capture asymmetries, multimodality, and varying uncertainty structure, while also generating the true cluster mass more accurately than the $M$–$\sigma$ estimate.

Furthermore, we incorporate physical knowledge into the architecture by predicting residual corrections to the standard $M$–$\sigma$ relation, rather than directly predicting the cluster mass. This design enhances the interpretability of our model, allowing us to explicitly isolate information beyond what is captured by equilibrium-based assumptions.

\begin{figure*}[!htb]
    \centering
    \includegraphics[width=0.95\linewidth]{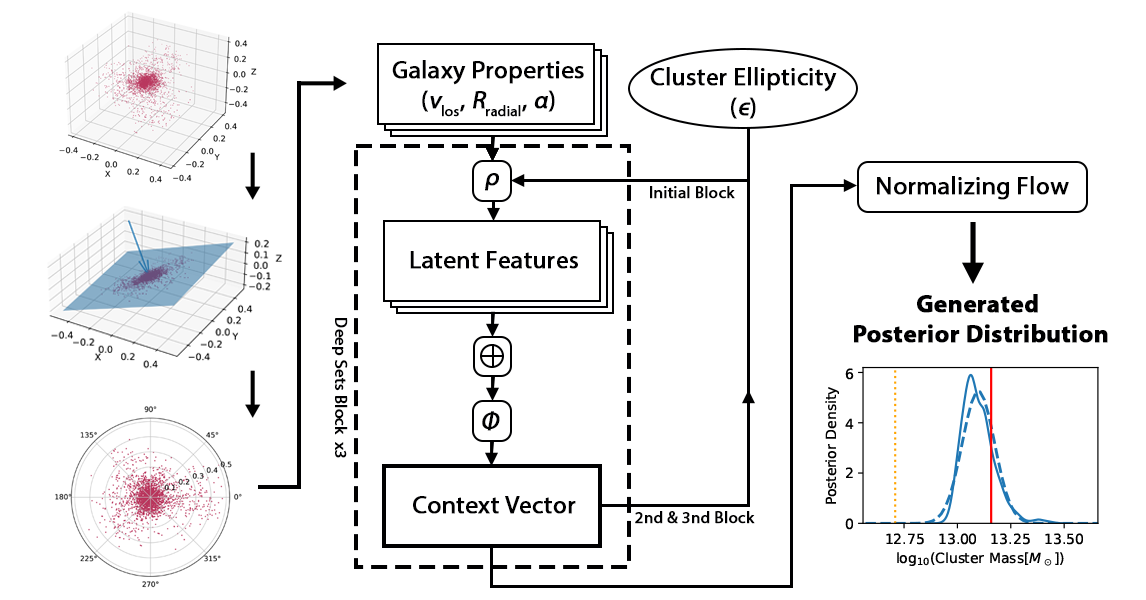}
    \caption{Schematic overview of our model for probabilistic cluster mass estimation. Galaxy member properties—line-of-sight velocity \(v_{\text{los}}\), projected radial distance \(R_{\text{radial}}\), and angular position \(\alpha\)—are obtained by projecting galaxies onto randomly selected planes. These features are processed through three Deep Sets blocks to produce permutation-invariant latent representations. The initial block incorporates a global cluster-level feature, ellipticity \(\epsilon\), which is updated and passed through subsequent blocks. The final latent vector serves as context for a normalizing flow, which outputs a full posterior distribution over cluster mass.}
    \label{fig:method}
\end{figure*}

\section{Method}
Here, we describe our dataset, model architecture and selected features to infer posterior distributions of galaxy cluster masses from member galaxy dynamics. The full implementation is available at our GitHub repository\footnote{https://github.com/BonnyWang/Cluster-Mass-DeepSet-NF}.

\subsection{Dataset}
To model realistic galaxy cluster dynamics, we use the Uchuu-UniverseMachine dataset \citep{Aung2023}, a large-volume $(2\, \text{Gpc}/h)^3$ $N$-body simulation with $12,\!800^3$ particles. We analyze galaxy clusters from the Rockstar finder \citep{Behroozi2013}, and galaxy properties from UniverseMachine \citep{Behroozi2019}. This provides robust and realistic representations of both galaxies and galaxy clusters. The baseline $M$-$\sigma$ relation for galaxy clusters in this dataset is
\begin{equation}
\log_{10}(\sigma_v) = 0.33 \log_{10}(M_{\text{vir}}) - 1.95 + \gamma,
\end{equation}
where \( M_{\text{vir}} \) is the virial mass, \( \sigma_v \) is the member galaxy velocity dispersion, and $\gamma$ is a random variable drawn from a distribution that depends on the mass range and other properties of the galaxy cluster. We split the data into 70\% for training, 15\% for validation, and 15\% for testing.

\subsection{Model Architecture}

To probe whether galaxy dynamics contain additional information beyond the standard \( M\text{-}\sigma \) relation, we train a machine learning model to predict the residual between the true cluster mass and a baseline $M$-$\sigma$ estimate. The corrected mass prediction is:

\begin{equation}
    \log_{10} M_{\text{vir}}^{\text{pred}} = \log_{10} M_{\text{vir}}^{M\text{-}\sigma} + \Delta_{\text{ML}},
\end{equation}

where \( \Delta_{\text{ML}} \) is a learned correction modeled probabilistically. Our architecture consists of two components: a stack of three Deep Sets blocks for feature extraction and a conditional normalizing flow for posterior generation.

Each Deep Sets block here consists of a pointwise encoder \( \rho \) and a context updater \( \phi \), both modeled as multilayer perceptrons (MLPs), along with a permutation-invariant aggregation \( \bigoplus \). Notably, we use mean pooling to avoid incorporating information related to galaxy richness. Given input galaxy features \( \vec{p}_i \) for member galaxies in a cluster, along with an initial global cluster-level feature vector \(\vec{u}^{(0)}\), the architecture performs the following operations:

\begin{align}
    \vec{h}_i^{(k)} &= \rho^{(k)}\left([\vec{p}_i, \vec{u}^{(k-1)}]\right), \quad 
    \vec{u}^{(k)} = \phi^{(k)}\left( \bigoplus \vec{h}_i^{(k)} \right), 
\end{align}
where \( \vec{h}_i^{(k)} \) is the latent feature of the \( i \)-th galaxy after the \( k \)-th encoder, and \( \vec{u}^{(k)} \) is the global context vector summarizing the cluster state at block \( k \).

As we stack three Deep Sets blocks, the final output \( \vec{u}^{(3)} \) serves as a latent context vector summarizing the internal galaxy dynamics of each cluster. This vector conditions a normalizing flow that models the residual \( \Delta_{\text{ML}} \) as a flexible, non-Gaussian distribution. We implement this using the conditional spline transformation provided by \textsc{Pyro} \citep{Bingham2018Pyro}. Finally, all components are jointly optimized by maximizing the conditional log-likelihood of the posterior:
\begin{equation}
    \mathcal{L} = -\log p(\Delta_{\text{ML}} \mid \vec{u}^{(3)}).
\end{equation}
The progression of training and validation loss values is shown in Appendix \ref{sec:appendix}.

In addition, to have an equal number of samples across all mass ranges, we augmented the dataset to enforce a uniform mass distribution across training samples.

\subsection{Observational Feature}

To mimic the observational process, we construct our dataset by first selecting a random line of sight and projecting member galaxies onto a plane perpendicular to it. To preserve the cluster’s overall morphology while ensuring rotational invariance, we fit an ellipse to the projected distribution and take its semi-major axis as a reference. Based on this setup, we select the following features as inputs to our machine learning model: Line-of-sight velocity (\( v_{\text{los}} \)), radial distance to the halo center (\( R_{\text{radial}} \)), angle relative to the fitted ellipse axis (\( \alpha \)). In addition, we include a global feature to characterize the galaxy distribution: the ellipticity (\( \epsilon \)), computed from the fitted ellipse’s semi-major and semi-minor axes.

These features serve as inputs to the machine learning framework described in the previous section. The full pipeline, including the construction of observational features and the machine learning architecture, is illustrated in Figure~\ref{fig:method}.

\begin{figure*}
    \centering
    \includegraphics[width=0.328\linewidth]{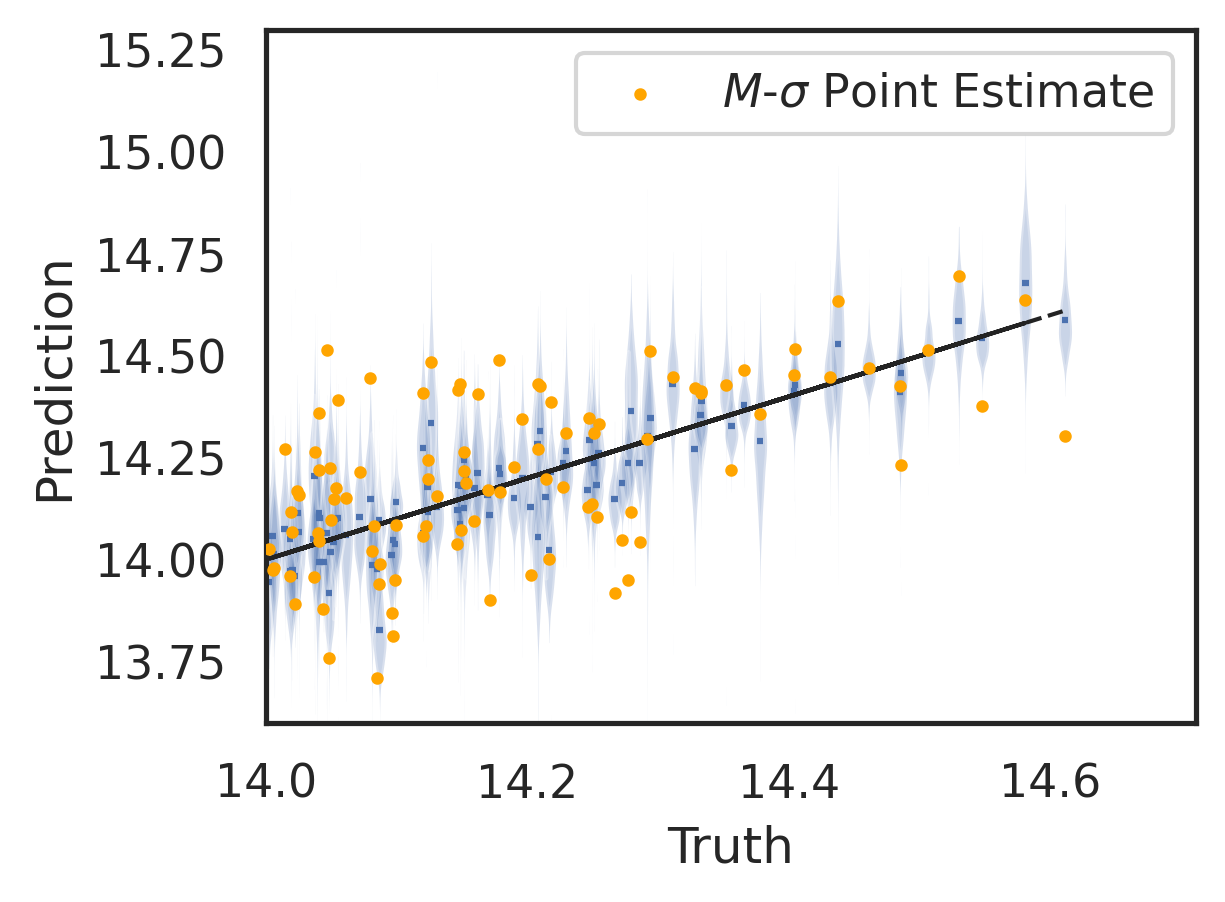}
    \includegraphics[width=0.328\linewidth]{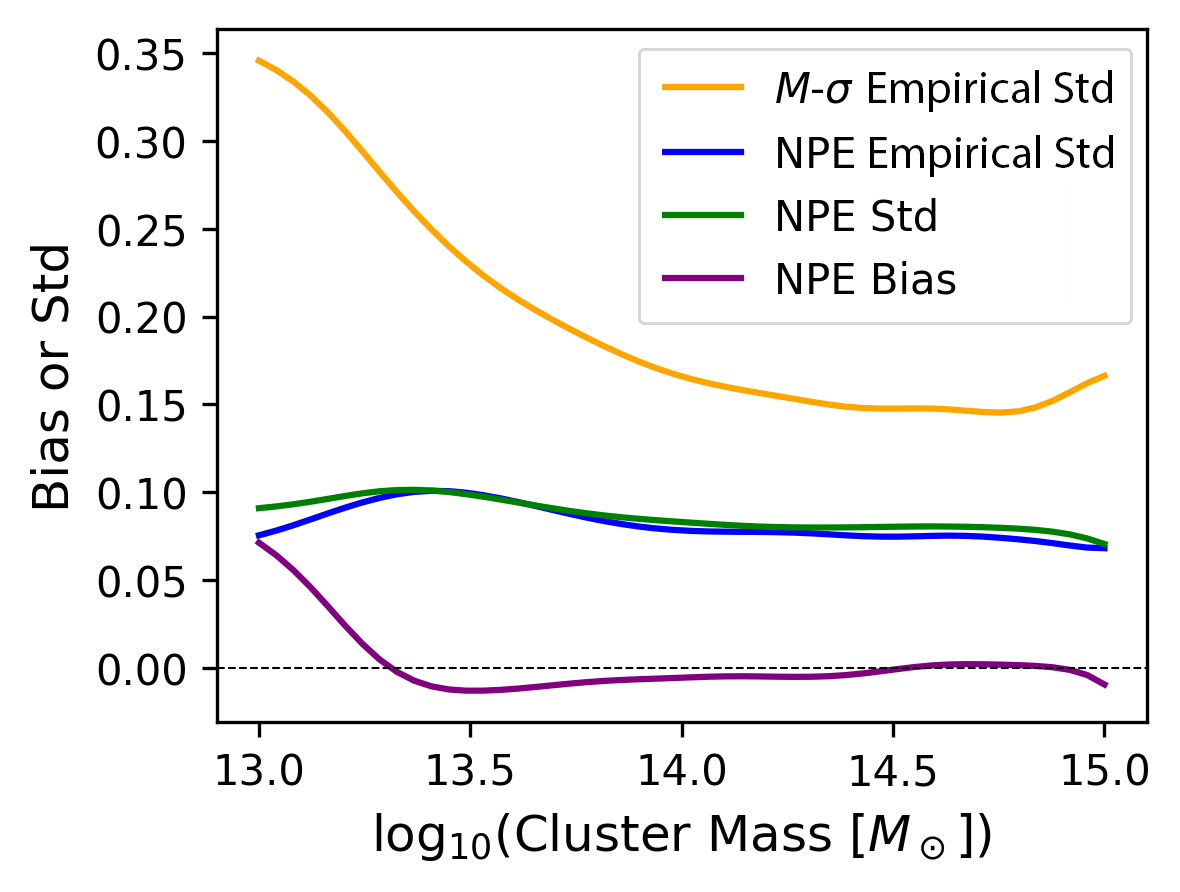}
    \includegraphics[width=0.328\linewidth]{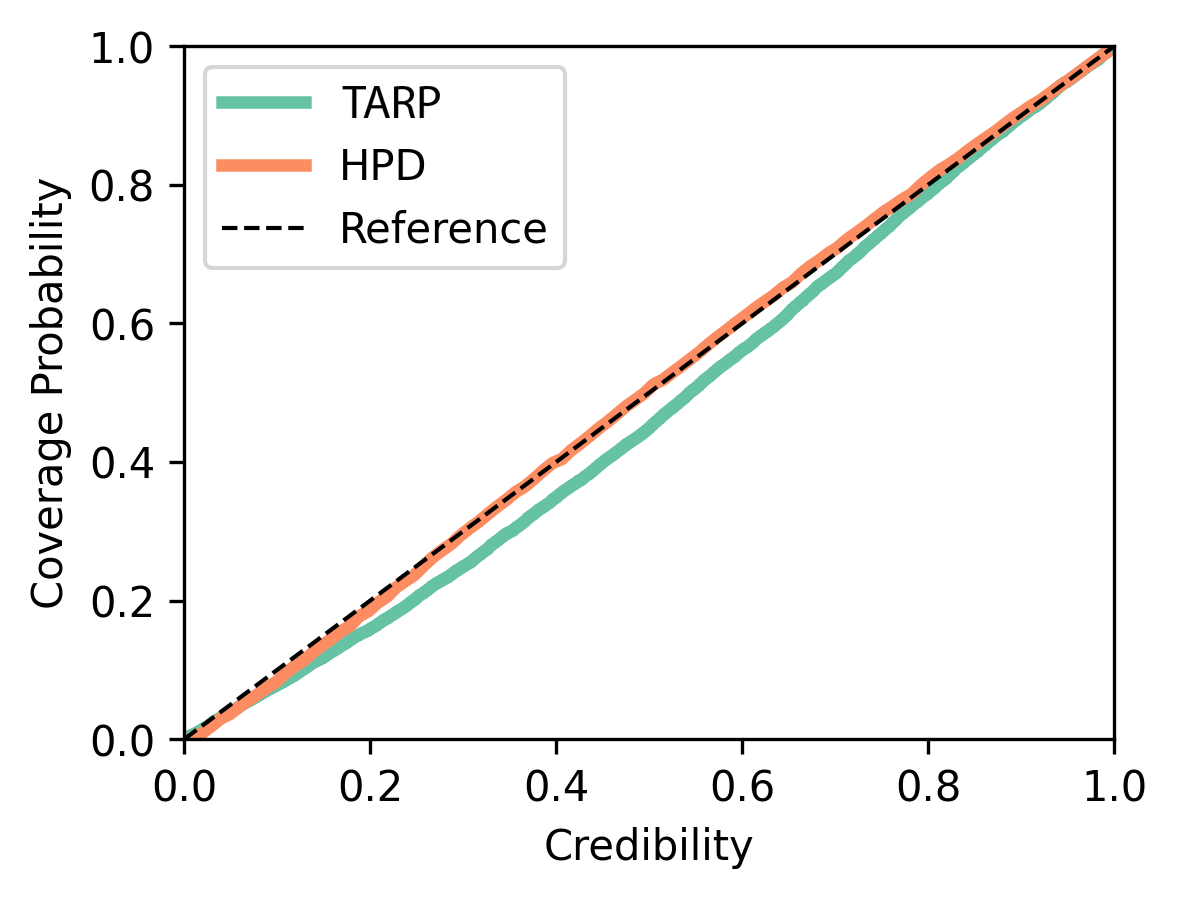}
    \caption{Left: Comparison between true and predicted galaxy cluster masses. The x-axis shows true masses; the y-axis shows predictions. Orange points represent baseline $M$-$\sigma$ estimates, and violin plots indicate the machine learning posterior distributions. The black diagonal denotes perfect agreement. Middle: Bias and standard deviation in cluster mass as a function of true mass. The orange and blue lines show the standard deviation of $M$-$\sigma$ and NPE mean relative to true mass, respectively; the green line shows the mean predicted posterior standard deviation; and the purple line shows the mean prediction bias. The dashed black line indicates zero bias. Right: Uncertainty calibration assessment using highest posterior density (HPD, orange) and Tests of Accuracy with Random Points (TARP, green).}
    \label{fig:main_results}
\end{figure*}
\begin{figure}[!htb]
    \centering
    \includegraphics[width=\linewidth]{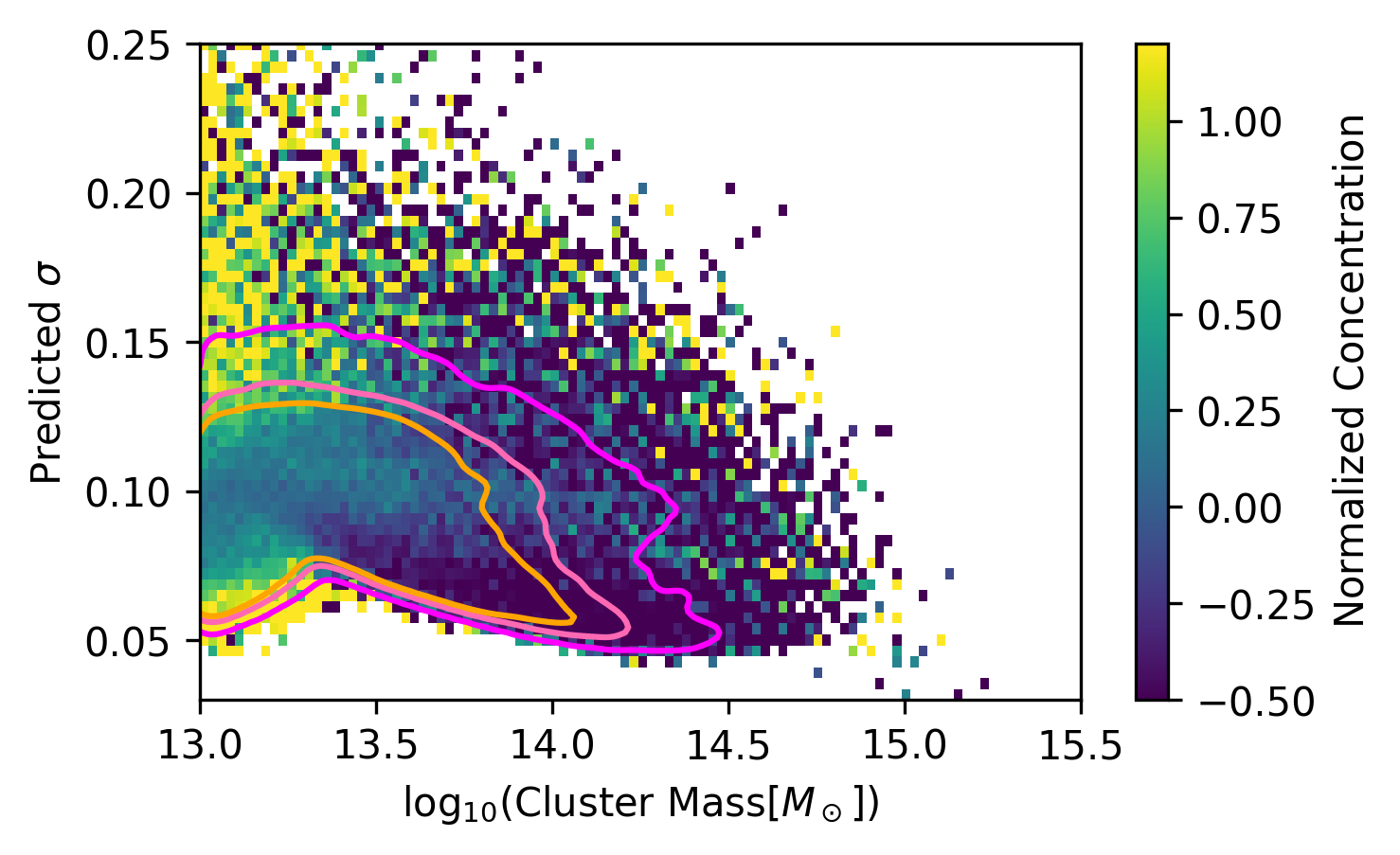}
    \caption{Predicted posterior standard deviation as a function of true cluster mass, color-coded by the normalized cluster concentration. The overlaid contours mark the 90\%, 94\%, and 98\% levels of the number density distribution, suggesting a bimodal pattern: the upper peak is associated with higher concentrations, while the lower peak corresponds to lower concentrations.}
    \label{fig:concentration}
\end{figure}
\section{Result}
In this section, we present the predictions from our machine learning framework, followed by an analysis of how galaxy cluster properties influence these predictions.

\subsection{Machine Learning Correction}
We show our main result comparing machine learning predicted and true galaxy cluster masses in the left panel of Figure \ref{fig:main_results}. True masses are on the x-axis; predicted masses are on the y-axis. Orange points indicate traditional $M$-$\sigma$ estimates, while violin plots visualize the model's posterior distributions. The black diagonal marks perfect agreement. Most posteriors are centered closer to this line than $M$-$\sigma$ predictions.

The middle panel of Figure~\ref{fig:main_results} illustrates the bias and standard deviation in cluster mass predictions as a function of true mass. The standard deviation of the NPE empirical scatter (blue) closely matches the predicted posterior standard deviation (green) across most mass ranges, indicating well-calibrated uncertainty estimates. Compared to the standard deviation of the $M$-$\sigma$ point estimate (orange), the NPE empirical standard deviation exhibits approximately twice the reduced scatter to the truth. This highlights a great improvement in mass accuracy. The model shows little bias across all masses, as evidenced by the near-zero purple line.

Moreover, we assess uncertainty calibration using two methods: the highest posterior density (HPD) and TARP~\citep[Tests of Accuracy with Random Points,][]{Lemos2023}. In the right panel of Figure \ref{fig:main_results}, both HPD (orange) and TARP (green) closely follow the black dashed identity line, again, indicating well-calibrated posteriors.

\subsection{Galaxy Property Analysis}
To gain further insight into the behavior of our machine learning model, we examine how other cluster properties relate to its predictions. We observe a noticeable impact of cluster concentration on the predicted posterior uncertainty.

In Figure~\ref{fig:concentration}, we show the predicted posterior standard deviation as a function of true cluster mass, color-coded by normalized cluster concentration. The normalization removes the mass dependence of concentration. The overlaid contours represent the number density distribution of clusters. We notice a bimodal structure in the contours: the upper peak corresponds to higher concentrations, while the lower peak aligns with lower concentrations. This suggests that the model yields tighter constraints for less concentrated, likely more unrelaxed, clusters. This contradicts our expectation that relaxed systems are easier to constrain under equilibrium assumptions. Several possible explanations exist: overconcentrated clusters may be in fact out of equilibrium or in a transient phase of their dynamical evolution \citep{Ludlow2012, Sereno2013}, and the fraction of dynamically unrelaxed clusters increases at the high-mass end \citep{Ludlow2013}.

\section{Conclusion and Discussion}
We present a machine learning framework that combines Deep Sets and conditional normalizing flows to infer full posterior distributions of galaxy cluster masses from projected galaxy dynamics. Using the Uchuu-UniverseMachine simulation and a physically motivated set of observational features, we explore residual corrections beyond the standard $M$–$\sigma$ relation. This residual information may reflect complex dynamical processes such as galaxy merger histories, motivating further investigation.

In terms of methodology, we explicitly incorporate projected positional information through a set-based architecture. This allows the model to learn directly from the full phase-space configuration of cluster member galaxies. By correcting the $M$-$\sigma$ mass estimate via learned residuals, our framework utilizes physical priors from analytic models and adds more interpretability to the prediction. Finally, the use of conditional normalizing flows enables full posterior estimation.

Compared to the baseline $M$-$\sigma$ estimates, our results significantly reduce scatter by approximately a factor of two and produce well-calibrated uncertainties across the full mass range. Notably, we find that the predicted uncertainty varies somewhat with cluster concentration: the model yields slightly tighter constraints for less concentrated and likely more dynamically unrelaxed clusters. While this trend deviates from expectations based on equilibrium assumptions, it may reflect subtle dynamical effects captured by the model. This finding also highlights the potential of machine learning not only to improve predictive performance but also to reveal nuanced patterns in cluster physics. Further investigation is needed to understand the origin of this concentration-dependent behavior and its implications for cluster physics.

Finally, for application to real observations such as DESI \citep{DESI2016}, PFS \citep{Tamura2016}, Rubin/LSST \citep{Ivezic2019}, and Euclid \citep{Laureijs2011}, a few additional steps are needed to test the robustness of our model. First, we need to evaluate its reliability with respect to membership selection, including completeness and interloper contamination. Second, we should assess its performance across different simulations, since it is currently trained within a specific cosmology. Although the highly nonlinear and largely virialized nature of cluster dynamics suggests limited sensitivity to cosmological variations, future work is necessary to confirm these effects.

\section*{Acknowledgements}
We thank Hy Trac, Matthew Ho, and Michelle Ntampaka for valuable discussions. We thank the reviewers for their valuable and constructive feedback, which helped improve the quality of our work.
This research used computing resources at Kavli IPMU and the Flatiron Institute.
The Kavli IPMU is supported by the WPI (World Premier International Research Center) Initiative of the MEXT (Japanese Ministry of Education, Culture, Sports, Science and Technology).
Leander Thiele was supported by JSPS KAKENHI Grant 24K22878.
\section*{Impact Statement}
This paper presents work whose goal is to advance the field of 
applying Machine Learning in astrophysical problems. There are many potential societal consequences 
of our work, none which we feel must be specifically highlighted here.

\nocite{langley00}

\bibliography{main}
\bibliographystyle{icml2025}

\newpage
\appendix
\onecolumn
\section{Training and validation loss curves}
\label{sec:appendix}
\begin{figure}[!htb]
    \centering
    \includegraphics[width=0.95\linewidth]{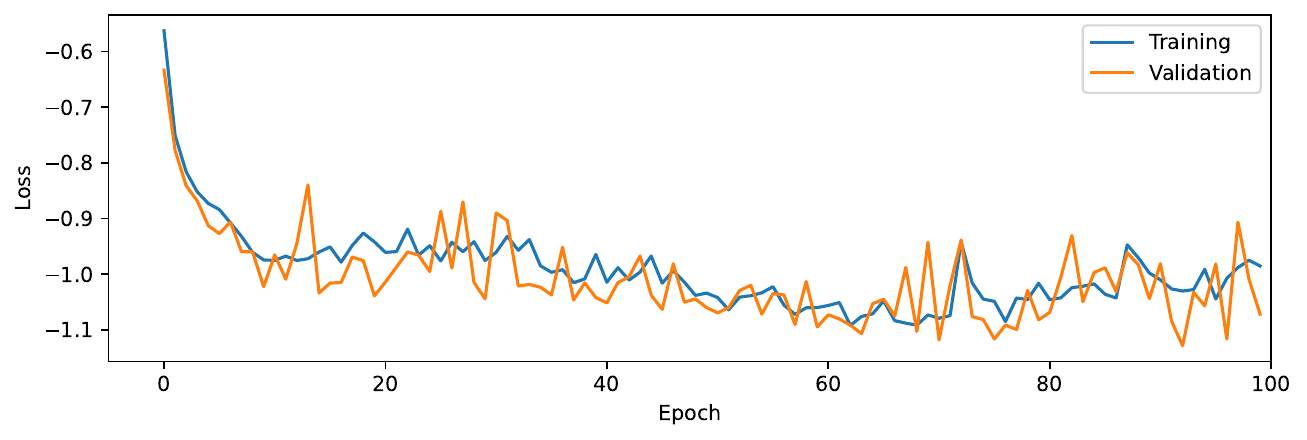}
    \caption{Conditional log-likelihood loss during model training. Both training and validation losses generally decrease over epochs, indicating improved model fit. }
    \label{fig:enter-label}
\end{figure}

\end{document}